# A Comparison of Axiomatic Approaches to Qualitative Decision Making Using Possibility Theory

Phan H. Giang and Prakash P. Shenoy

University of Kansas School of Business, Summerfield Hall
Lawrence, KS 66045-2003, USA
*phgiang@ku.edu, pshenoy@ku.edu*

## Abstract

In this paper we analyze two recent axiomatic approaches proposed by Dubois *et al.* [5] and by Giang and Shenoy [10] for qualitative decision making where uncertainty is described by possibility theory. Both axiomtizations are inspired by von Neumann and Morgenstern's system of axioms for the case of probability theory. We show that our approach naturally unifies two axiomatic systems that correspond, respectively, to pessimistic and optimistic decision criteria proposed by Dubois *et al.* The simplifying unification is achieved by (*i*) replacing axioms that are supposed to reflect two informational attitudes (uncertainty aversion and uncertainty attraction) by an axiom that imposes order on set of standard lotteries, and (*ii*) using a binary utility scale in which each utility level is represented by a pair of numbers.

## 1 Introduction

The main goal of this paper is to compare two recent axiomatic approaches proposed by Dubois *et al.* [5] and Giang and Shenoy [10] for qualitative decision making where uncertainty is described by possibility theory.

In recent years, there is growing interest in qualitative decision making within the AI community [3, 2]. The aim of the research is to deal with various situations where probability and utility inputs required by the classical decision theory are difficult to assess. It is long recognized that probability theory can not faithfully capture all facets of uncertainty that is pervasive in the world. Among several alternative approaches proposed in AI to deal with uncertainty, belief function theory [15, 16], interval-valued probability [13, 20] and fuzzy possibility theory [21, 6] occupy prominent positions. Once uncertainty has been represented, the next step is to determine how it can be used in decision making. For the first two theories where probabilistic semantics are still relevant, a standard solution is to assess (according to some criteria) a probability distribution and then apply the classical decision theory. For possibility theory, which apparently has no such strong connection with probability, the technique is of little use. In recent years, efforts have been made to create an axiomatic basis for decision theory tailored for possibility theory. Stylistically, the efforts are in two different but related directions following von Neumann - Morgenstern and Savage [5, 9].

This paper is structured as follows. In the next section, the proposal by Dubois *et al.* for a decision theory with possibility theory is reviewed. In section 3, we will present a new system of axioms that has been modified from our previous proposal designed for Spohnian epistemic belief theory. We prove a representation theorem for that system. In the section 4, a comparison between two approaches is made. We prove a theorem stating that the two axiomatic systems by Dubois *et al.* are just special cases of our system. An example that illustrates calculation with different utility functions is provided. The last section consists of some concluding remarks.

## 2 Pessimistic and Optimistic Utilities

In this section, we review, with some terminology modification, the proposal that has been exposed in a series of papers by Dubois *et al.* [8, 4, 5]. Assume a set $S$ of possible *situations* or *states*. A finite uncertainty scale $V$ is assumed, without loss of generality, to be a set of points in the unit interval $[0, 1]$ such that $0, 1 \in V$. Order $\geq$ on $V$ is defined in a natural way. Uncertainty about which among possible states will occur is captured by a possibility distribution that is a mapping $\pi : S \to V$ such that $\max_{s \in S} \pi(s) = 1$. The possibility of a subset $A \subseteq S$, $\pi(A) \stackrel{\text{def}}{=} \max_{s \in A} \pi(s)$. A finite set $X = \{x_1, x_2, \ldots x_n\}$ of *consequences* or *outcomes* or *prizes* is also given. To avoid triviality, $X$ is assumed to have at least two elements ($n \geq 2$). We also assume the existence of two distinct *anchor*[1] elements in set $X$. $\overline{x}$ is the *best* and $\underline{x}$ is the *worst* i.e., $\overline{x} \succeq x_i$ and $x_i \succeq \underline{x}$ $\forall i$ where $\succeq$ is a preference relation with the reading "at least as good as". A *decision* or *lottery* is a mapping from $S \to X$. That is, decision $d$ delivers outcome $d(s)$ in the case that state $s$ occurs. Notice that each decision $d$ induces a possibility distribution $\pi_d$ on the set of consequences in the following sense: $\pi_d(x) \stackrel{\text{def}}{=} \pi(d^{-1}(x))$ where $d^{-1}(x) = \{s \in S | d(s) = x\}$. Denote by $\Pi_X$ the set of possibility distributions on $X$. Set $\Pi_X$ is closed under an operation (possibilistic) *mix*. For $\pi_1, \pi_2 \in \Pi_X$ and $\lambda, \mu \in [0,1]$ such that $\max(\lambda, \mu) = 1$, a *mixture of $\pi_1, \pi_2$ with weights* $\lambda, \mu$ denoted by $(\lambda/\pi_1, \mu/\pi_2)$ can be defined as follows $(\lambda/\pi_1, \mu/\pi_2)(x) = \max(\min(\lambda, \pi_1(x)), \min(\mu, \pi_2(x)))$. Mixture operation is pairwise commutative in the sense that $(\lambda/\pi_1, \mu/\pi_2) = (\mu/\pi_2, \lambda/\pi_1)$. It is useful to consider a generalized version of mix operation: a mixture of $m$ possibility functions $\pi_1, \ldots \pi_m$ with weights $\lambda_1, \ldots \lambda_m$ such that $\max_{1 \leq i \leq m}\{\lambda_i\} = 1$ defined as

$$(\lambda_1/\pi_1, \ldots \lambda_m/\pi_m)(x) \stackrel{\text{def}}{=} \max_{1 \leq i \leq m} \min(\lambda_i, \pi_i(x)). \quad (1)$$

To a decision maker, the value of decision $d$ is the same as the value of the induced $\pi_d$. And thanks to the mixture construct, set $\Pi_X$ is rich enough to encode not only a simple decision but also a "compound" decision. For an intuitive reference, a reader can find a similarity between the concept of possibilistic mixture and that of two-stage lottery or "randomized" decisions in probabilistic approach. A logical conclusion of the above argument is that an analysis of preference on decisions boils down to the analysis of set $\Pi_X$ of possibility distributions on the set of consequences $X$. In other words, preference of a decision maker can be analyzed through a preference relation $\succeq$ on $\Pi_X$. A preference relation could be characterized by a system of axioms (properties) it must satisfy or it could be modeled by a utility function that maps elements of $\Pi_X$ into some (finite) linearly ordered scale $U$ called the *utility scale*. $\sup(U) = 1$ and $\inf(U) = 0$ are assumed. Symbol $\geq$ is used for both numerical and utility comparison.

Dubois *et al.* consider two kinds of utility called respectively *pessimistic* and *optimistic* utility and propose two axiomatic systems to characterize them.

### 2.1 Pessimistic Utility

The pessimistic utility concept needs the following ingredients: a function

$$u : X \to U \quad (2)$$

that determines utility for each outcome such that $u(\overline{x}) = 1$ and $u(\underline{x}) = 0$; a function

$$n : U \to U \quad (3)$$

that is an order reversing involution in $U$ i.e. $n(1) = 0$, $n(0) = 1$ and $n(u_1) \geq n(u_2)$ whenever $u_1 \leq u_2$; and a function

$$h : V \to U \quad (4)$$

that is an order preserving mapping from uncertainty scale $V$ onto utility scale $U$ such that $h(1) = 1$ and $h(0) = 0$. Given that, a *pessimistic qualitative utility* function $QU^- : \Pi_X \to U$ is defined as

$$QU^-(\pi) \stackrel{\text{def}}{=} \min_{x \in X} \max(nh(\pi(x)), u(x)) \quad (5)$$

where $nh \stackrel{\text{def}}{=} n \circ h$ – a composition of $n$ and $h$.

Dubois *et al.* also show the following equality

$$QU^-((\lambda/\pi_1, \mu/\pi_2)) = \min \left\{ \begin{array}{l} \max(nh(\lambda), QU^-(\pi_1)) \\ \max(nh(\mu), QU^-(\pi_2)) \end{array} \right\} \quad (6)$$

Given a utility function, one can specify a preference relation [2] $\succeq$ on $\Pi_X$. Alternatively, that relation can be characterized by the following axiom system denoted by $\mathcal{S}_P$

$A1^-$ (Total pre-order) $\succeq$ is reflexive, transitive and complete.

$A2^-$ (Uncertainty aversion) If $\pi \leq \pi'$ then $\pi \succeq \pi'$.

$A3^-$ (Subsitutability) If $\pi_1 \sim \pi_2$ then $(\lambda/\pi_1, \mu/\pi) \sim (\lambda/\pi_2, \mu/\pi)$.

$A4^-$ (Continuity) $\forall \pi \in \Pi_X, \exists \lambda \in V \;\; \pi \sim (1/\overline{x}, \lambda/\underline{x})$.

Dubois *et al.* prove the following representation theorem

**Theorem 1** *A preference relation $\succeq$ on $\Pi_X$ satisfies system $\mathcal{S}_P$ iff there exist functions $u, n, h$ and $QU^-$ defined as by (2, 3, 4, 5) such that $\pi \succeq \pi'$ iff $QU^-(\pi) \geq QU^-(\pi')$.*

---

[1] The term is chosen to reflect the role these elements play in the continuity axiom presented later in the section.

[2] We also use two derivative relations: $\succ$ for strict preference and $\sim$ for indifference.

## 2.2 Optimistic Utility

The authors also consider another utility that supposedly captures the optimistic behavior of decision makers. The optimistic qualitative utility function $QU^+ : \Pi_X \to U$ is defined as follows

$$QU^+(\pi) \stackrel{\text{def}}{=} \max_{x \in X} \min(h(\pi(x)), u(x)) \quad (7)$$

The system $S_O$ of axioms that characterize $QU^+$ is obtained from $S_P$ by replacing axioms $A2^-, A4^-$ by $A2^+$ and $A4^+$ respectively where

$A2^+$ (Uncertainty attraction) If $\pi \geq \pi'$ then $\pi \succeq \pi'$.

$A4^+$ (Continuity) $\forall \pi \in \Pi_X, \exists \lambda \in V \ \pi \sim (\lambda/\overline{x}, 1/\underline{x})$.

They also prove a representation theorem for $QU^+$ and $S_O$ which is similar to Theorem 1.

## 3 Unified Possibilistic Utility

In this section, we will translate the construct of qualitative utility [10] that was originally proposed for Spohn's theory of epistemic belief into the possibility theory framework.

A theory of epistemic belief, originally proposed by Spohn [17, 18] to deal with plain belief, has its roots in Adams's [1] work on the logic of conditionals. Spohn's theory has been studied extensively by Goldszmidt and Pearl [11, 12] under the name "rank-based system" or "qualitative probabilities" or "$\kappa$-calculus". The basic construct of the theory is the concept of disbelief function $\delta : S \to \mathcal{N}$ such that $\min_{s \in S} \delta(s) = 0$ where $\mathcal{N}$ is the set of non-negative integers. For $A \subseteq S$, $\delta(A) \stackrel{\text{def}}{=} \min_{s \in A} \delta(s)$. For $A \subseteq S$ and $s \in A$, the conditional disbelief function $\delta(s|A)$ is defined as $\delta(s|A) \stackrel{\text{def}}{=} \delta(\omega) - \delta(A)$. Despite some nuances, there is a tight relationship between possibility theory and Spohn's theory through log-transformation that has been pointed out in [7]. Namely, for a disbelief function $\delta$, $c^{-\delta}$ is a possibility function where $c > 1$ is a constant. Conversely, if $\pi$ is a possibility function then $Int[-\log_c(\pi)]$ is a disbelief function where $Int[.]$ is a integer extracting function.

Here are some technical notes. In [10], we have 6 axioms that were inspired by presentation of von Neumann and Morgernstern's axiom system by Luce and Raiffa [14]. Here, in order to make later comparison more transparent, we will present those axioms in a slightly modified form. In [10] we used the product-based mixture in the set of Spohnian lotteries in this paper we will adopt the min-based mixture for possibilistic lotteries.

We use term *standard* lottery for a lottery that realizes (with corresponding degrees of certainty) in either the best prize $\overline{x}$ or the worst $\underline{x}$, i.e., $(\lambda/\overline{x}, \mu/\underline{x})$ where $\lambda, \mu \in V$ and $\max(\lambda, \mu) = 1$. We use **B** to denote the set of all standard lotteries.

We have the following system of axioms that is denoted by $S$ without subscript.

$B1$ (Total pre-order) $\succeq$ is reflexive, transitive and complete.

$B2$ (Qualitative monotonicity) $\succeq$ restricted over **B** satisfies the following condition. Let suppose $\sigma = (\lambda/\overline{x}, \mu/\underline{x})$ and $\sigma' = (\lambda'/\overline{x}, \mu'/\underline{x})$ then

$$\sigma \succeq \sigma' \quad \text{iff} \quad \begin{cases} 1 \geq \lambda \geq \lambda' \ \& \ \mu = \mu' = 1 \\ \lambda = 1 \ \& \ \lambda' < 1 \\ \lambda = \lambda' = 1 \ \& \ \mu \leq \mu' \end{cases} \quad (8)$$

$B3$ (Subsitutability) If $\pi_1 \sim \pi_2$ then $(\lambda/\pi_1, \mu/\pi) \sim (\lambda/\pi_2, \mu/\pi)$.

$B4$ (Continuity) $\forall x \in X, \exists \sigma \in \mathbf{B} \ \ x \sim \sigma$.

We list the axioms of $S$ in the same order as those of $S_P$. Compared with the system in [10], we note the following correspondence: $B1$ (Total pre-order) axiom incorporates Axioms 1 and 5 (order of prizes and transitivity), $B2$ is Axiom 6, $B3$ is Axiom 3 and $B4$ is Axiom 4. Reduction of compound lotteries axiom is taken care of by the definition of possibilistic mixture.

We need a lemma.

**Lemma 1** *Assume $\succeq$ satisfies $S$ (axioms $B1$ through $B4$). For each $\pi \in \Pi_X$, there exists one and only one $\sigma \in \mathbf{B}$ such that $\pi \sim \sigma$.*

**Proof:** By definition of possibilistic mixture (1), $\pi$ can be rewritten in the form of a mixture $(\pi(x_1)/x_1, \pi(x_2)/x_2, \ldots \pi(x_n)/x_n)$. By $B4$, we have $x_i \sim \sigma_i$ for $1 \leq i \leq n$ where $\sigma_i$ is a standard lottery $\sigma_i = (\lambda_i/\overline{x}, \mu_i/\underline{x})$. By $B3$, we have $\pi \sim (\pi(x_1)/\sigma_1, \pi(x_2)/\sigma_2, \ldots \pi(x_n)/\sigma_n)$. Again using the definition of mixture, we have $(\pi(x_1)/\sigma_1, \pi(x_2)/\sigma_2, \ldots \pi(x_n)/\sigma_n) = (\lambda/\overline{x}, \mu/\underline{x})$ where

$$\lambda = \max_{1 \leq i \leq n} \min(\pi(x_i), \lambda_i)$$

$$\mu = \max_{1 \leq i \leq n} \min(\pi(x_i), \mu_i)$$

So $\pi \sim (\lambda/\overline{x}, \mu/\underline{x})$. By $B1$ and $B2$, $(\lambda/\overline{x}, \mu/\underline{x})$ must be unique. ∎

Let us consider a utility function $QU : \Pi_X \to U$. If we wish that $\pi_1 \succeq \pi_2$ iff $QU(\pi_1) \geq QU(\pi_2)$ holds, from qualitative monotonicity ($B2$), it is clear that utility

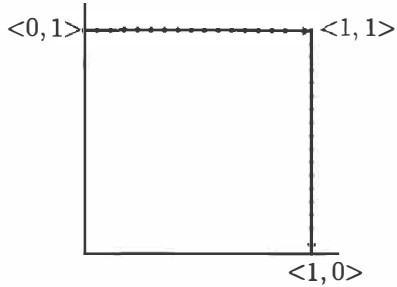

Figure 1: Binary utility scale $U_V$.

scale $U$ must be at least rich enough to distinguish every standard lottery. Let us take

$$U_V \stackrel{\text{def}}{=} \{<\lambda,\mu> \mid \lambda,\mu \in V \text{ and } \max(\lambda,\mu) = 1\}. \quad (9)$$

In other words, $U_V$ is the set of pair of elements in the uncertainty scale $V$ such that one of them is 1. A linear order $\geq$ on $U_V$ is defined as, for $u = <\lambda,\mu>$, $u' = <\lambda',\mu'>$,

$$u \geq u' \text{ iff } \begin{cases} 1 \geq \lambda \geq \lambda' \ \& \ \mu = \mu' = 1 \\ \lambda = 1 \ \& \ \lambda' < 1 \\ \lambda = \lambda' = 1 \ \& \ \mu \leq \mu' \end{cases} \quad (10)$$

We refer to $U_V$ equipped with the order $\geq$ as the *binary utility scale*.

We extend[3] operation $min$ in such a way that it is distributive with respect to pairing as follows

$$\min(\alpha,<\beta,\gamma>) \stackrel{\text{def}}{=} <\min(\alpha,\beta),\min(\alpha,\gamma)> \quad (11)$$

We also extend operation $max$ so that it is associative with respect to pairing

$$\max(<\alpha,\beta>,<\gamma,\delta>) \stackrel{\text{def}}{=} <\max(\alpha,\gamma),\max(\beta,\delta)> \quad (12)$$

Let us call a function $u : X \to U_V$ that assigns utility for each prize in $X$ a *basic utility assessment*. We say that a basic utility assessment is *consistent* with $\succeq$ if for any pair $x, y \in X$ $x \succeq y$ iff $u(x) \geq u(y)$, $u(\overline{x}) = <1,0>$ and $u(\underline{x}) = <0,1>$. Remember that $\overline{x}$ and $\underline{x}$ are respectively the best and the worst prizes in $X$. For a given basic utility assessment $u$, let us define a utility function $QU : \Pi_X \to U_V$ as follows

$$QU(\pi) \stackrel{\text{def}}{=} \max_{x \in X} \min(\pi(x), u(x)) \quad (13)$$

We have the following lemma on standard lotteries

**Lemma 2** *Suppose* $\sigma = (\lambda/\overline{x}, \mu/\underline{x})$ *is a standard lottery then* $QU(\sigma) = <\lambda,\mu>$.

---
[3]We decide in favor of extending operations $min$ and $max$ instead of creating new symbols. Hopefully, this slight abuse of notation does not lead to any confusion because the type of arguments will tell which rule is to apply.

**Proof:** We have by definition of $QU$

$$\begin{aligned} QU(\sigma) &= \max(\min(\lambda,<1,0>),\min(\mu,<0,1>)) \\ &= \max(<\lambda,0>,<0,\mu>) \\ &= <\lambda,\mu>. \ \blacksquare \end{aligned}$$

Now we have the following representation theorem

**Theorem 2** $\succeq$ *on* $\Pi_X$ *satisfies axioms* $B1$ *through* $B4$ *iff there exists a consistent basic utility assessment* $u$ *such that* $\pi \succeq \pi'$ *iff* $QU(\pi) \geq QU(\pi') \ \forall \pi, \pi' \in \Pi_X$ *where* $QU$ *is defined by (13).*

**Proof:**

($\Rightarrow$) Suppose $\succeq$ satisfies axioms $B1$ through $B4$. For $\pi_1, \pi_2 \in \Pi_X$, let us assume $\pi_1(x_i) = \pi_{1i}$ and $\pi_2(x_i) = \pi_{2i}$ for $1 \leq i \leq n$. Suppose $\pi_1 \succeq \pi_2$, we will show that $QU(\pi_1) \geq QU(\pi_2)$.

By $B4$, for each $x_i \in X$ we have $x_i \sim \sigma_i$ for some standard lottery $\sigma_i = (\lambda_i/\overline{x}, \mu_i/\underline{x})$. Let us select a function $u$ as follows $u(\overline{x}) = <1,0>$, $u(\underline{x}) = <0,1>$ and

$$u(x_i) = <\lambda_i, \mu_i> \text{ for } 1 \leq i \leq n. \quad (14)$$

By $B3$, $\pi_i = (\pi_{i1}/x_1, \pi_{i2}/x_2, \ldots \pi_{in}/x_n) \sim (\pi_{i1}/\sigma_1, \pi_{i2}/\sigma_2, \ldots \pi_{in}/\sigma_n)$ for $i = 1, 2$. Let us give a name $\rho_i$ to the right hand sides ($i = 1, 2$). We will show that $\rho_i$, which is a possibilistic mixture, is reduced to a standard lottery. By definition of mixture,

$$\rho_i(\overline{x}) = \max_{1 \leq j \leq n} \min(\pi_{ij}, \lambda_j) \quad (15)$$

$$\rho_i(\underline{x}) = \max_{1 \leq j \leq n} \min(\pi_{ij}, \mu_j) \quad (16)$$

and for all other $x \in X$

$$\rho_i(x) = \max_{1 \leq j \leq n} \min(\pi_{ij}, 0) = 0 \quad (17)$$

By $B1$, from $\pi_1 \succeq \pi_2$ we have $\rho_1 \succeq \rho_2$. By $B2$, since $\rho_i$ are standard lotteries, $\rho_1 \succeq \rho_2$ means either

$$\begin{aligned} &(\rho_1(\overline{x}) \geq \rho_2(\overline{x}) \text{ and } \rho_1(\underline{x}) = \rho_2(\underline{x}) = 1) \text{ or} \\ &(\rho_1(\overline{x}) = 1 \text{ and } \rho_2(\overline{x}) \leq 1) \text{ or} \\ &(\rho_1(\overline{x}) = \rho_2(\overline{x}) = 1 \text{ and } \rho_1(\underline{x}) \leq \rho_2(\underline{x})). \end{aligned} \quad (18)$$

Now, let us consider the pair $<\rho_i(\overline{x}), \rho_i(\underline{x})>$. By equations (15, 16) and definitions (11, 12), we have

$$<\rho_i(\overline{x}), \rho_i(\underline{x})> = \max_{1 \leq j \leq n} \min(\pi_{ij}, <\lambda_j, \mu_j>) \quad (19)$$

Taking into account equations (19, 14, 13), we have $QU(\pi_i) = <\rho_i(\overline{x}), \rho_i(\underline{x})>$. And each of conditions (18) implies $QU(\pi_1) \geq QU(\pi_2)$.

($\Leftarrow$) For a given $u : X \to U_V$ such that $u(\overline{x}) = <1,0>$ and $u(\underline{x}) = <0,1>$, a function $QU : \Pi_X \to U_V$ is defined as in (13). We have to show that order $\succeq$ on $\Pi_X$ induced by $QU$ ($\pi \succeq \pi'$ iff $QU(\pi) \geq QU(\pi')$) satisfies axioms $B1$ through $B4$.

$B1$ is satisfied because the order on $U_V$ is transitive and complete.

Suppose $\sigma \succeq \sigma'$ where $\sigma = (\lambda/\overline{x}, \mu/\underline{x})$, $\sigma' = (\lambda'/\overline{x}, \mu'/\underline{x})$ are two standard lotteries. Because $\succeq$ is induced from $QU$, we have $QU(\sigma) \geq QU(\sigma')$. By lemma 2 we have $<\lambda, \mu> \geq <\lambda', \mu'>$. By definitions (10) and (8) we infer that $\succeq$ satisfies $B2$.

Suppose $\pi_1 \sim \pi_2$. Because $\succeq$ is induced from $QU$, we have $QU(\pi_1) = QU(\pi_2)$. By definition (13), we will have

$$QU((\lambda/\pi_1, \mu/\pi)) =$$
$$= \max(\min(\lambda, QU(\pi_1)), \min(\mu, QU(\pi)))$$
$$= \max(\min(\lambda, QU(\pi_2)), \min(\mu, QU(\pi)))$$
$$= QU((\lambda/\pi_2, \mu/\pi))$$

This means $\succeq$ induced by $QU$ satisfies $B3$.

Finally, the existence of basic utility assessment $u$ together with lemma 2 guarantee satisfaction of $B4$. ∎

## 4  Pessimistic, Optimistic or Unified Utilities

In this section, we will do a comparison of three systems of qualitative utilities presented in previous sections. Since we have a representation theorem for each of them, we can discuss the systems either in terms of axioms or in terms of utility functions $QU^-$, $QU^+$, and $QU$.

First of all, note that the adjectives "pessimistic" and "optimistic" used for axiom systems $\mathcal{S}_P, \mathcal{S}_O$, perhaps, implicitly refer to the opposite direction of axiom $A2^-$ and $A2^+$. If $\pi_1 \geq \pi_2$ in numerical sense i.e. $\pi_1(x_i) \geq \pi_2(x_i)$ $\forall i$, then $A2^-$ requires $\pi_2 \succeq \pi_1$ while $A2^+$ requires $\pi_1 \succeq \pi_2$. $A2^-$ and $A2^+$ are called uncertainty attitude axioms. The former is "uncertainty aversion" and the latter is "uncertainty attraction".

But perhaps these names are a source of confusion. First, since axiom systems $\mathcal{S}_P, \mathcal{S}_O$ are presented in style of von Neumann and Morgenstern [19], it is appropriate to recall similar terms "risk aversion" and "risk attraction." In the linear utility theory, risk aversion (attraction) refers to the concavity (convexity) of utility function. In other words, risk aversion and risk attraction are properties ascribed to individual utility functions. They are not a property of the utility theory. Different psychological states may result in different utility assignment to the same outcome or a different assessment of uncertainty related to outcomes, but it is hard to conceive that they require different theories as implied by $\mathcal{S}_P$ and $\mathcal{S}_O$. Operationally, the dichotomy of pessimistic and optimistic systems might also lead to difficulty in application. For example, how would a decision maker classify herself as either "optimistic" or "pessimistic" or what would happen if she was unsure about either options. Moreover, in the writing of the authors for example [8, 6], it is clear that inequality of the form $\pi_1 \geq \pi_2$ is an informational relationship. It says that $\pi_2$ is more *specific* than $\pi_1$. In other words, it says that $\pi_2$ contains more information than $\pi_1$ does. So it seems to us, the equation of informational relationship with preferential relationship $\succeq$ is not a very sensible idea. Although information has its own value, informational value *per se* rarely serves as a decision criterion. For example, decision making under uncertainty is mostly guided by von Neumann and Morgenstern's linear utility theory rather than by Shannon's information theory.

Let us consider the following example. We face a choice between two lotteries $\pi_1 = (1/\overline{x}, 1/\underline{x})$ and $\pi_2 = \underline{x}$. In other words, $\pi_1$ is a possibilistic distribution on $X$ such that $\pi_1(\overline{x}) = \pi_1(\underline{x}) = 1$, $\pi_1(x) = 0$ for all other $x$ and $\pi_2(\underline{x}) = 1$, $\pi_2(x) = 0$ for all other $x$. According to possibility theory [8, 6], $\pi_1$ describes a situation where we have knowledge to exclude all prizes except $\overline{x}$ and $\underline{x}$. Moreover, we are equally sure about occurence of either of the prizes. $\pi_2$ describes a *complete knowledge* situation where all but $\underline{x}$ are excluded. Because something is going to happen, $\pi_2$ is equivalent to saying that $\underline{x}$ is the certain prize. Axiom $A2^-$ will force us to consider[4] $\pi_2$ is *at least* as good as $\pi_1$. In other words, if we were adopting $\mathcal{S}_P$ we would have been indifferent between a surely worst prize and an uncertain outcome in which there is a hope to get the best prize. We believe such a choice is unreasonable. To see when axiom $A2^+$ recommends a bad action, we could consider a choice between a surely best prize and uncertain outcome where there is a danger of getting the worst prize. $A2^+$ will recommend the latter. Note that all these anomalies are corrected by axiom $B2$.

Let us consider sets of standard lotteries $\mathbf{B} = \{(\lambda/\overline{x}, \mu/\underline{x}) | \lambda, \mu \in V$ and $\max(\lambda, \mu) = 1\}$, $\mathbf{B}^- \stackrel{\text{def}}{=} \{(1/\overline{x}, \mu/\underline{x}) | \mu \in V\}$ and $\mathbf{B}^+ \stackrel{\text{def}}{=} \{(\lambda/\overline{x}, 1/\underline{x}) | \lambda \in V\}$. We have $\mathbf{B} = \mathbf{B}^- \cup \mathbf{B}^+$ and $\mathbf{B}^- \cap \mathbf{B}^+ = (1/\overline{x}, 1/\underline{x})$. It is straightforward to verify the following lemma

**Lemma 3** *Let $\succeq^-, \succeq^+$ and $\succeq$ denote respectively the order relations on $\mathbf{B}^-$, $\mathbf{B}^+$ and $\mathbf{B}$ imposed by $A2^-$, $A2^+$ and $B2$, $\succeq = \succeq^- \cup \succeq^+ \cup (\mathbf{B}^- \times \mathbf{B}^+)$*

---

[4]To be exact, a complete calculation would show that $\pi_1 \sim \pi_2$. But a *lazy evaluation* would suggest to choose $\pi_2$ over $\pi_1$ without bothering further calculation.

The lemma implies that $\succeq^-$ is the same as $\succeq$ restricted to $\mathbf{B}^-$ and $\succeq^+$ is the same as $\succeq$ restricted to $\mathbf{B}^+$.

It is easy to check that $QU^-((\lambda/\overline{x}, 1/\underline{x})) = QU^-(\underline{x})$, $QU^+((1/\overline{x}, \mu/\underline{x})) = QU^+(\overline{x})$. In general $QU^-$ and $QU^+$ have the following property

**Lemma 4** *Suppose $QU^-(\pi_1) \geq QU^-(\pi_2)$ then*

$$QU^-((\lambda/\pi_1, 1/\pi_2)) = QU^-(\pi_2) \quad (20)$$
$$QU^+((1/\pi_1, \mu/\pi_2)) = QU^+(\pi_1). \quad (21)$$

**Proof:** We will prove (20). The proof of (21) is just dually similar. Using equation (6), we have

$$QU^-((\lambda/\pi_1, 1/\pi_2)) = \min \left\{ \begin{array}{l} \max(nh(\lambda), QU^-(\pi_1)) \\ \max(nh(1), QU^-(\pi_2)) \end{array} \right\} \quad (22)$$

By definition of function $nh$ (eqs. 3,4), $nh(1) = 0$. So, $\max(nh(1), QU^-(\pi_2)) = QU^-(\pi_2)$. Because of lemma condition $QU^-(\pi_1) \geq QU^-(\pi_2)$, $\max(nh(\lambda), QU^-(\pi_1)) \geq QU^-(\pi_2)$. By equation (22), we have $QU^-((\lambda/\pi_1, 1/\pi_2)) = QU^-(\pi_2)$. ∎

Thus, if a decision maker is pessimistic, whenever she sees that the less desirable prize of a lottery is fully possible she will ignore all considerations about other prizes and uncertainty to conclude that the lottery is as good as the least desirable prize. For a optimistic decision maker, once she sees the more desirable prize of a lottery is fully possible she concludes that the lottery is worth the same as that best prize. Roughly speaking, $QU^-$ ($QU^+$) lumps together a half of total number of lotteries.

It has been noted that $QU^-$ and $QU^+$ are "complementary" in a sense that although optimistic $QU^+$ is not able to distinguish two lotteries $(1/\overline{x}, \mu_1/\underline{x})$ and $(1/\overline{x}, \mu_2/\underline{x})$, pessimistic $QU^-$ can discriminate between them by comparing values of $\mu_1$ and $\mu_2$. The situation is reversed for lotteries $(\lambda_1/\overline{x}, 1/\underline{x})$ and $(\lambda_2/\overline{x}, 1/\underline{x})$. Again, we note that $QU$ agrees with $QU^-$ in the former situation and with the $QU^+$ in the latter situation.

With notations $\mathbf{B}^-$ and $\mathbf{B}^+$, axioms $A4^-$ and $A4^+$ can be restated respectively as $\forall \pi \in \Pi_X, \exists \sigma \in \mathbf{B}^- \ \pi \sim \sigma$ and $\forall \pi \in \Pi_X, \exists \sigma \in \mathbf{B}^+ \ \pi \sim \sigma$. We'll show that these axioms can be weakened, without any effect to the results, by requiring instead $\forall x \in X$. i.e.,

$(B4^-) \ \forall x \in X, \ \exists \sigma \in \mathbf{B}^- \ x \sim \sigma$.

$(B4^+) \ \forall x \in X, \ \exists \sigma \in \mathbf{B}^+ \ x \sim \sigma$.

As we argued previously, these axioms are counterintuitive. $A4^-$ requires, for example, the worst prize $\underline{x}$ is equivalent to some lottery where the best prize $\overline{x}$ is fully possible. But the presense of $A2^-$ ($A2^+$) makes the stated form of $A4^-$ ($A4^+$) necessary. Had $A4^-$ been substituted by $B4 \ \forall x \in X, \exists \sigma \in \mathbf{B} \ x \sim \sigma$, we would still have $\underline{x} \sim (1/\overline{x}, 1/\underline{x})$, because $(1/\overline{x}, 1/\underline{x})$ was the mimimal element in $\mathbf{B}$ according to $A2^-$. Let assume for some $x' \succ \underline{x}$ ($x'$ is strictly preferred to $\underline{x}$) $x' \sim (\lambda/\overline{x}, 1/\underline{x})$. Using definition (5) and the facts that $nh(1) = 0$, $u(\overline{x}) = 1$ and $u(\underline{x}) = 0$, we calculate $QU^-((\lambda/\overline{x}, 1/\underline{x})) = \min(\max(nh(\lambda), u(\overline{x})), \max(nh(1), u(\underline{x}))) = 0$. From that we infer $QU^-(x') = 0 = QU^-(\underline{x})$. This is inconsistent with assumption $x' \succ \underline{x}$.

We have the following theorem that states precisely the relationship between systems $\mathcal{S}_P$, $\mathcal{S}_O$ and $\mathcal{S}$.

**Theorem 3**

$(i) \quad \mathcal{S}_P \models \mathcal{S}$
$(ii) \quad \mathcal{S}_O \models \mathcal{S}$
$(iii) \quad \mathcal{S} \cup B4^- \models \mathcal{S}_P$
$(iv) \quad \mathcal{S} \cup B4^+ \models \mathcal{S}_O$

**Proof:** We will prove $(i)$ and $(iii)$. The proof of $(ii)$ and $(iv)$ is dually similar.

$(i)$ Assume $A1^-$ through $A4^-$ are satisfied, since $B1$ is the same as $A1^-$ and $B3$ is the same as $A3^-$, we are left to prove that $B2$ and $B4$ are also satisfied. From $A4^-$ for each $\pi \in \Pi_X, \exists \sigma \in \mathbf{B}^- \ \pi \sim \sigma$. Obviously, $X \subseteq \Pi_X$ and $\mathbf{B}^- \subseteq \mathbf{B}$, so $B4$ is also satisfied (note that symbol $X$ is used for the set of prizes as well as the set of singleton possibility distributions on set of prizes). And finally, we will show [5] that $\mathcal{S}_P \models B2$. Assume that $\succeq$ satisfies $\mathcal{S}_P$. For two standard lotteries $\sigma = (\lambda/\overline{x}, \mu/\underline{x})$ and $\sigma' = (\lambda'/\overline{x}, \mu'/\underline{x})$. We want to show

$$\sigma \succeq \sigma' \text{ iff } \left\{ \begin{array}{ll} 1 \geq \lambda \geq \lambda' \ \& \ \mu = \mu' = 1 & (1) \\ \lambda = 1 \ \& \ \lambda' < 1 & (2) \\ \lambda = \lambda' = 1 \ \& \ \mu \leq \mu' & (3) \end{array} \right. \quad (23)$$

(If) Observe that if $\mu = 1$ ($\mu' = 1$), by theorem 1 and lemma 4, we have $\sigma \sim \underline{x} \sim (1/\overline{x}, 1/\underline{x})$ ($\sigma' \sim \underline{x} \sim (1/\overline{x}, 1/\underline{x})$). In case (1) when $\mu = \mu' = 1$, we have $\sigma \sim \underline{x} \sim \sigma'$. Thus, $\sigma \succeq \sigma'$. In case (2), since $\max(\lambda', \mu') = 1$ from $\lambda' < 1$ we have $\mu' = 1$. Therefore, $\sigma' \sim \underline{x} \sim (1/\overline{x}, 1/\underline{x})$. By axiom $A2^-$ we have $\sigma \succeq (1/\overline{x}, 1/\underline{x})$. From transitivity, $\sigma \succeq \sigma'$. In case (3), since $\lambda = \lambda' = 1$ and $\mu \leq \mu'$, by $A2^-$ we have $\sigma \succeq \sigma'$.

(Only If) Assume $\sigma \succeq \sigma'$. By $A4^-$ we can assume $\lambda = \lambda' = 1$. Furthermore, $\mu > \mu'$ would violate $A2^-$. We have $\mu \leq \mu'$. Thus, the right hand side of (23) (a disjunction) is true.

$(iii)$ Note that since $B4^- \models B4$, set of axioms $\mathcal{S} \cup B4^-$ is effectively one that is obtained by replacing $B4$ by

---

[5] Note that $A2^-$ states only a sufficient condition for $\succeq$ while $B2$ states both necessary and sufficient conditions.

$B4^-$. We have to show that if $B1, B2, B3$ and $B4^-$ are satisfied so are $A1^-, A2^-, A3^-$ and $A4^-$. Again, we do not have to worry about $A1^-$ and $A3^-$ since they are identical to $B1$ and $B3$.

First, we will show the satisfaction of $A4^-$. From $B4^-$, we can assume $x_i \sim \sigma_i$ for $1 \leq i \leq n$ where $\sigma_i \in \mathbf{B}^-$. We will show that $\forall \pi \in \Pi_X, \exists \sigma \in \mathbf{B}^- \; \pi \sim \sigma$. Suppose $\sigma_i = (1/\overline{x}, \mu_i/\underline{x})$ for $1 \leq i \leq n$. By $B3$, $\pi \sim (\pi(x_1)/\sigma_1, \pi(x_2)/\sigma_2 \ldots \pi(x_n)/\sigma_n)$. Applying the definition of mixture (1) for the right hand side, say $\rho$, of the indifference,

$$\rho(\overline{x}) = \max_{1 \leq i \leq n} \min(\pi(x_i), 1) \qquad (24)$$

$$\rho(\underline{x}) = \max_{1 \leq i \leq n} \min(\pi(x_i), \mu_i) \qquad (25)$$

and for all other $x \in X$

$$\rho(x) = \max_{1 \leq i \leq n} \min(\pi(x_i), 0) = 0 \qquad (26)$$

We have $\rho(\overline{x}) = 1$ because $\max_i(\pi(x_i)) = 1$. Thus $\pi \sim (1/\overline{x}, \rho(\underline{x})/\underline{x})$.

Now we turn to $A2^-$. Suppose $\pi_1 \geq \pi_2$. We just show that $\pi_1 \sim (1/\overline{x}, \rho_1(\underline{x})/\underline{x})$ and $\pi_2 \sim (1/\overline{x}, \rho_2(\underline{x})/\underline{x})$ where $\rho_i(\underline{x})$ $i = 1, 2$ is calculated by equation (25). Since $\pi_1 \geq \pi_2$, we have $\rho_1(\underline{x}) \geq \rho_2(\underline{x})$. By $B2$, we have $(1/\overline{x}, \rho_2(\underline{x})/\underline{x}) \succeq (1/\overline{x}, \rho_1(\underline{x})/\underline{x})$. From this, by transitivity we have $\pi_2 \succeq \pi_1$. ∎

**Corollary 1** *System $\mathcal{S}_P$ ($\mathcal{S}_O$) is a special case of $\mathcal{S}$ when each prize in $X$ has an equivalent standard lottery in $\mathbf{B}^-$ ($\mathbf{B}^+$).*

Since $\mathcal{S}_P$ and $\mathcal{S}_O$ are special cases of $\mathcal{S}$, a question that can be raised [6] is if a "combination" of those special cases has the same expressive power as $\mathcal{S}$. Specifically, if one considers pairs of pessimistic and optimistic utilities of lotteries $<QU^-(\pi), QU^+(\pi)>$ can one come to something similar to binary utility $QU(\pi)$? We suspect the answer is no.[7]

**Example:** Let $X = \{x_1, x_2, x_3, x_4\}$. $\overline{x} = x_1 \succ x_2 \succ x_3 \succ x_4 = \underline{x}$. $V = \{1, .7, .5, 0\}$ and $U = \{1, .5, .3, 0\}$. Consider $\pi_1, \pi_2 \in \Pi_X$ with $\pi_1(x_1) = .7$, $\pi_1(x_2) = 1$, $\pi_1(x_3) = .5$, $\pi_1(x_4) = .5$ and $\pi_2(x_1) = 1$, $\pi_2(x_2) = .7$, $\pi_2(x_3) = 0$, $\pi_2(x_4) = 1$. We will compare utility of $\pi_1$ and $\pi_2$ using $QU^-$ and $QU$.

For definition of $QU^-$, let us assume that function $n$ is given by $n(1) = 0$, $n(.5) = .3$, $n(.3) = .5$, $n(0) = 1$. Function $h$ is given by $h(1) = 1$, $h(.7) = .5$, $h(.5) = .3$, $h(0) = 0$. Their composition $nh$ is given by $nh(1) = 0$, $nh(.7) = .3$, $nh(.5) = .5$, $nh(0) = 1$. A utility assignment $u$ which is consistent with the preference order $x_1, x_2, x_3, x_4$ is given by $u(x_1) = 1$, $u(x_2) = .5$, $u(x_3) = .3$, $u(x_4) = 0$. Note that this utility assignment also means $x_1 \sim (1/x_1, 0/x_4)$, $x_2 \sim (1/x_1, .5/x_4)$, $x_3 \sim (1/x_1, .7/x_4)$ and $x_4 \sim (1/x_1, 1/x_4)$. Using definition (5) we calculate utility for $\pi_1$

$$QU^-(\pi_1) = \min \left\{ \begin{array}{c} \max(nh(.7), 1) \\ \max(nh(1), .5) \\ \max(nh(.5), .3) \\ \max(nh(.5), 0) \end{array} \right\}$$

$$= \min \left\{ \begin{array}{c} \max(.3, 1) \\ \max(0, .5) \\ \max(.5, .3) \\ \max(.5, 0) \end{array} \right\}$$

$$= \min\{1, .5, .5, .5\} = .5.$$

and $\pi_2$

$$QU^-(\pi_2) = \min \left\{ \begin{array}{c} \max(nh(1), 1) \\ \max(nh(.7), .5) \\ \max(nh(0), .3) \\ \max(nh(1), 0) \end{array} \right\}$$

$$= \min \left\{ \begin{array}{c} \max(0, 1) \\ \max(.3, .5) \\ \max(1, .3) \\ \max(0, 0) \end{array} \right\}$$

$$= \min\{1, .5, 1, 0\} = 0.$$

Thus, according to $QU^-$, $\pi_1$ which is equivalent to $x_2$, is strictly prefered to $\pi_2$, which is equivalent to $x_4$.

To define $QU$, we have $U_V = \{<0, 1>, <.5, 1>, <.7, 1>, <1, 1>, <1, .7>, <1, .5>, <1, 0>\}$. Take the following consistent basic utility assessment $u(x_1) = <1, 0>$, $u(x_2) = <1, .5>$, $u(x_3) = <1, .7>$ and $u(x_4) = <1, 1>$. Using definition (13) we calculate utility for $\pi_1$

$$QU(\pi_1) = \max \left\{ \begin{array}{c} \min(.7, <1, 0>) \\ \min(1, <1, .5>) \\ \min(.5, <1, .7>) \\ \min(.5, <1, 1>) \end{array} \right\}$$

$$= \max \left\{ \begin{array}{c} <.7, 0> \\ <1, .5> \\ <.5, .5> \\ <.5, .5> \end{array} \right\}$$

$$= <1, .5>$$

---
[6] It was raised by a referee.

[7] Here are some reasons for that. First, although $B4^- \models B4$ and $B4^+ \models B4$ we have $B4 \not\models B4^- \vee B4^+$. In other words, when a basic utility assignment equates elements in $X$ to standard lotteries on both halves of $U_V$, it violates both $A4^-$ and $A4^+$. In order that $<QU^-(\pi), QU^+(\pi)>$ makes sense, somehow at least one of them must hold. Second, on one hand, the set of $<QU^-(\pi), QU^+(\pi)>$ is a true two-dimensional object *i.e.*, there is no visible dependence between $QU^-(\pi)$ and $QU^+(\pi)$. On the other hand, set $U_V$ is not because one of the two numerical values in a pair must be 1.

and $\pi_2$

$$QU(\pi_2) = \max \left\{ \begin{array}{l} \min(1, <1,0>) \\ \min(.7, <1,.5>) \\ \min(0, <1,.7>) \\ \min(1, <1,1>) \end{array} \right\}$$

$$= \max \left\{ \begin{array}{l} <1,0> \\ <.7,.5> \\ <0,0> \\ <1,1> \end{array} \right\}$$

$$= <1,1>$$

So, according to $QU$, $\pi_1$ which is equivalent to $x_2$, is strictly prefered to $\pi_2$, which is equivalent to $x_4$. ∎

Let us consider the forms of functions $QU^-$, $QU^+$ and $QU$ given by equations (5), (7) and (13) respectively. First of all, it is easy to note that $QU$ looks similar to "optimistic" $QU^+$ which, in turn, is quite different from "pessimistic" $QU^-$. The form of $QU^+$ and $QU$ reminds us of the expected utility in probabilistic approach where the expected utility of a probabilistic lottery $p$ is defined as $EU(p) \stackrel{\text{def}}{=} \sum_{x \in X} p(x).u(x)$. Operations $max, min$ have, respectively, counterparts in addition $(+)$ and multiplication $(.)$. This similarity leads us to refer to $QU$ also as *expected qualitative utility function*. The difference between functions $QU^+$ and $QU$ is that latter makes no use of function $h$ that maps uncertainty scale $V$ onto utility scale $U$. In addition to $h$, definition of $QU^-$ requires an order reversing involution $n$ on $U$.

But the key distinction between $QU$ on one hand and $QU^-, QU^+$ on the other hand is that the utility scale used for $QU$ is an ordered set $U_V$ of *pairs* of numbers that are conveniently taken to be in the uncertainty scale $V$. The utility scale $U_V$ is chosen so that there is a one to one correspondence between $U_V$ and the set of standard lotteries **B**.

It is well known that probability theory interprets negation operation in a strictly complementary sense i.e., $p(A) = 1 - p(\neg A)$. That fact makes it sufficient to represent the occurence likelihood (or belief in Savage's personalistic view) of an event by one number - its probability. Unlike probability theory, possibility theory as well as most non-probabilistic uncertainty formalisms e.g., Demster-Shafer belief function theory [15, 16] or interval valued probabilty [13, 20] describe uncertainty of an event by *two* numbers. They are *possibility* and *necessity* degrees in possibility theory; *plausibility* and *belief* in DS theory; *upper* and *lower* probabilities in interval-valued probability theory. It is also well known that the heart of decision making under uncertainty is trading-off between uncertainty and utility. In order to enable the trade-off, utility and uncertainty must be "comparable." Therefore, binary utility is, perhaps, the "right" answer to binary uncertainty.

Focusing on standard lotteries, we can give the order on binary utility the following intuition. Since standard lottery $\sigma \in \mathbf{B}$ is a possibility distribution on $X$ such that $\sigma(\overline{x}) = \lambda, \sigma(\underline{x}) = \mu$ and $\sigma(x) = 0$ for all other $x \in X$, the possibility degree and the necessity degree assigned by $\sigma$ to $\overline{x}$ are $\lambda$ and $1 - \mu$. Because different standard lotteries have exactly the same prize, comparison of their utility boils down to comparing how sure the prize will be realized. The highest confidence level is, of course, represented by necessity degree 1 that corresponds to $\lambda = 1$ and $\mu = 0$. The confidence level is decreasing when necessity decreasing to 0. That corresponds to $\mu$ increases to 1. Before necessity degree becomes 0, the possibility is always 1. Once necessity equals 0, the confidence level can drop further with the falling of possibility degree from 1 to 0. The least confidence level is when the possibility degree is 0 i.e. $\lambda = 0$ and $\mu = 1$.

## 5 Conclusion

In this paper we have proposed a system of axioms for decision making with possibility theory. Our axiomatic system $(\mathcal{S})$ unifies the pessimistic and optimistic systems of axioms $(\mathcal{S}_P, \mathcal{S}_O)$ previously proposed by Dubois *et al*. The unification is made by $(i)$ replacing two informational attitude axioms $A2^-$ and $A2^+$ (uncertainty aversion and uncertainty attraction) by the monotonicity axiom $B2$; $(ii)$ generalization of continuity axioms $A4^-$ and $A4^+$ to axiom $B4$. Our axiom system subsumes both pessimistic and optimistic systems in the sense that any conclusion drawn by either $\mathcal{S}_P$ or $\mathcal{S}_O$ can also be made by $\mathcal{S}$. But the reverse is not true. Our system can sensibly handle situations where neither $\mathcal{S}_P$ nor $\mathcal{S}_O$ could. An example is when prizes in $X$ have equivalent standard lotteries in both halves $\mathbf{B}^-$ and $\mathbf{B}^+$ of $\mathbf{B}$. Beside the simplifying effect,[8] we argue that our proposal also removes uncertainty attitude from an utility theory to where it belongs – individual utility assessments.

We also prove a representation theorem for the unified system of axioms. Our utility function maps possibilistic lotteries into an ordered binary utility scale where each utility level is a pair of numbers. The utility function is a composition of $max, min$ operations that have been generalized in a natural way to work with pairs. The composition is similar to the composition of the classic expected utility expression where in place of $max$ is addition and in place of $min$ is multiplication. We also provide intuitive argument for the use

---

[8] Among other things, $QU$ does not need auxiliary functions $n$ and $h$ required by $QU^-$ and $QU^+$.

of binary utility.

### Acknowledgements

We thank Didier Dubois and anonymous referees for their useful suggestions and critiques to improve this paper. The first author is supported for this work by a grant from KU School of Business PhD Research Fund.